\pdfoutput=1
\documentclass[11pt]{article}
\usepackage{EMNLP2023}
\usepackage{times, latexsym, microtype, inconsolata, booktabs, url, hyperref, graphicx, array}
\usepackage{enumitem}
\usepackage[T1]{fontenc}
\usepackage[utf8]{inputenc}

\usepackage{amssymb}
\usepackage{pifont}
\usepackage{multirow}

\newcommand{\cmark}{\ding{51}}
\newcommand{\xmark}{\ding{55}}

\newcommand{\platform}{\texttt{Thresh}}
\newcommand{\pythonlib}{\texttt{thresh}}
\newcommand{\ntranslations}{14}
\newcommand{\ntutorials}{10}
\newcommand{\ntypologies}{11}

\newcommand{\logo}{\raisebox{-0.03\height}{\includegraphics[width=1.05em]{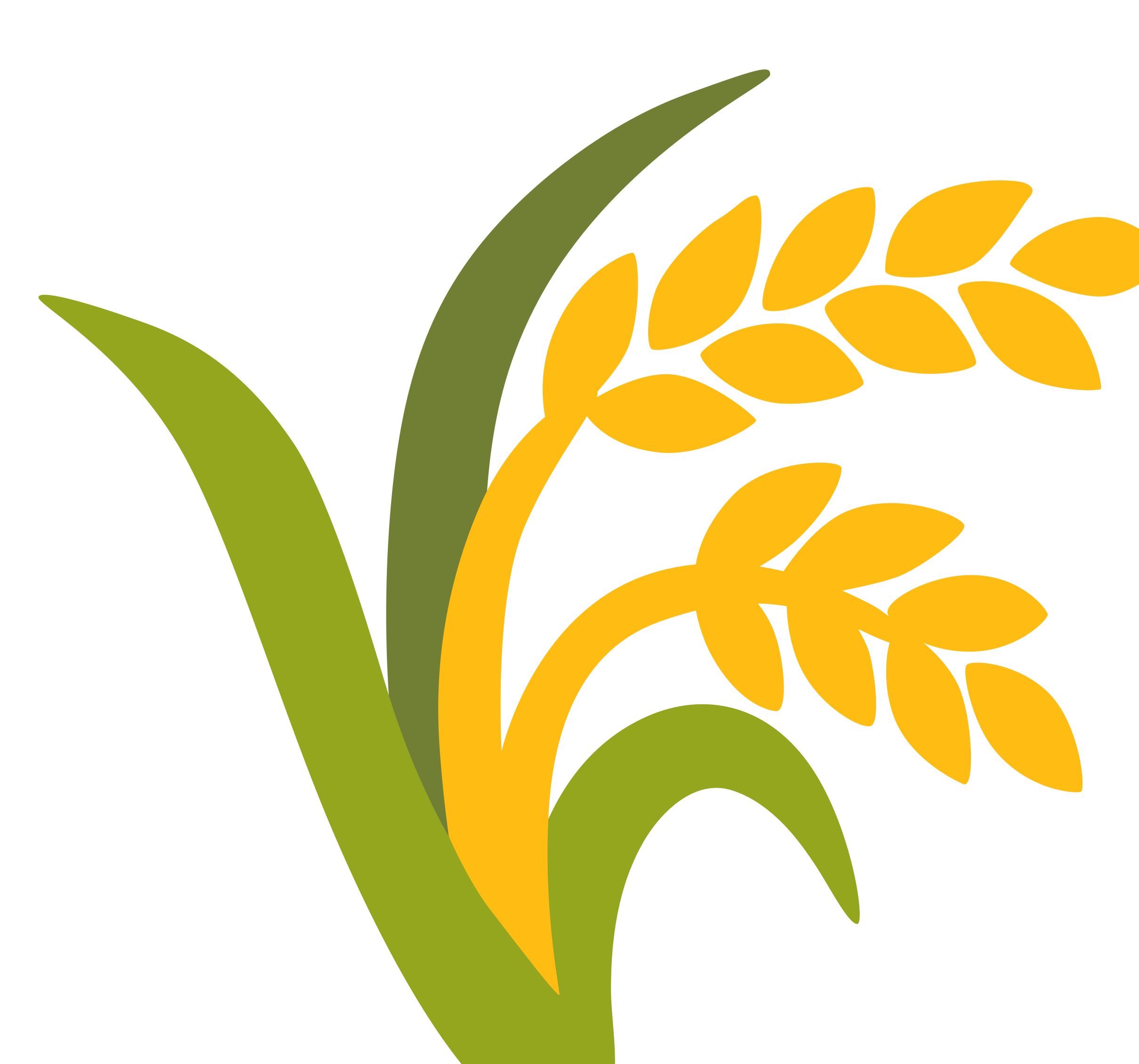}}}

\title{\platform{} \logo{}: A Unified, Customizable and Deployable Platform\\for Fine-Grained Text Evaluation}

\author{David Heineman, Yao Dou, Wei Xu \\
School of Interactive Computing, Georgia Institute of Technology \\
  {\small \texttt{\{david.heineman, douy\}@gatech.edu; wei.xu@cc.gatech.edu}} \\}
  
\begin{document}
\maketitle
\begin{abstract}
Fine-grained, span-level human evaluation has emerged as a reliable and robust method for evaluating text generation tasks such as summarization, simplification, machine translation and news generation, and the derived annotations have been useful for training automatic metrics and improving language models.
However, existing annotation tools implemented for these evaluation frameworks lack the adaptability to be extended to different domains or languages, or modify annotation settings according to user needs; and, the absence of a unified annotated data format inhibits the research in multi-task learning.
In this paper, we introduce \platform{} \logo{}, a unified, customizable and deployable platform for fine-grained evaluation. With a single YAML configuration file, users can build and test an annotation interface for any framework within minutes -- all in one web browser window.
To facilitate collaboration and sharing, \platform{} provides a community hub that hosts a collection of fine-grained frameworks and corresponding annotations made and collected by the community, covering a wide range of NLP tasks.
For deployment, \platform{} offers multiple options for any scale of annotation projects from small manual inspections to large crowdsourcing ones.
Additionally, we introduce a Python library to streamline the entire process from typology design and deployment to annotation processing.
\platform{} is publicly accessible at \url{https://thresh.tools}.
\end{abstract}

\section{Introduction}
\label{sec:introduction}
As modern large language models are able to generate human-level quality text \cite{Brown2020LanguageMA,OpenAI2023GPT4TR}, the evaluation of these models becomes increasingly challenging. Recent work has shown traditional surface-level evaluation methods such as pairwise comparison or Likert-scale ratings become less reliable \citep{clark-etal-2021-thats,maddela-etal-2023-lens} due to the close performance of these LLMs.
To address this, several fine-grained human evaluation frameworks have been proposed for various tasks such as open-ended generation \citep{dou-etal-2022-gpt}, text simplification \cite{heineman2023dancing}, and machine translation \cite{freitag-etal-2021-experts}. In these frameworks, annotators identify and annotate specific spans corresponding to quality or errors in the generated text.

\begin{figure}[t!]
    \centering
    \includegraphics[width=0.48\textwidth]{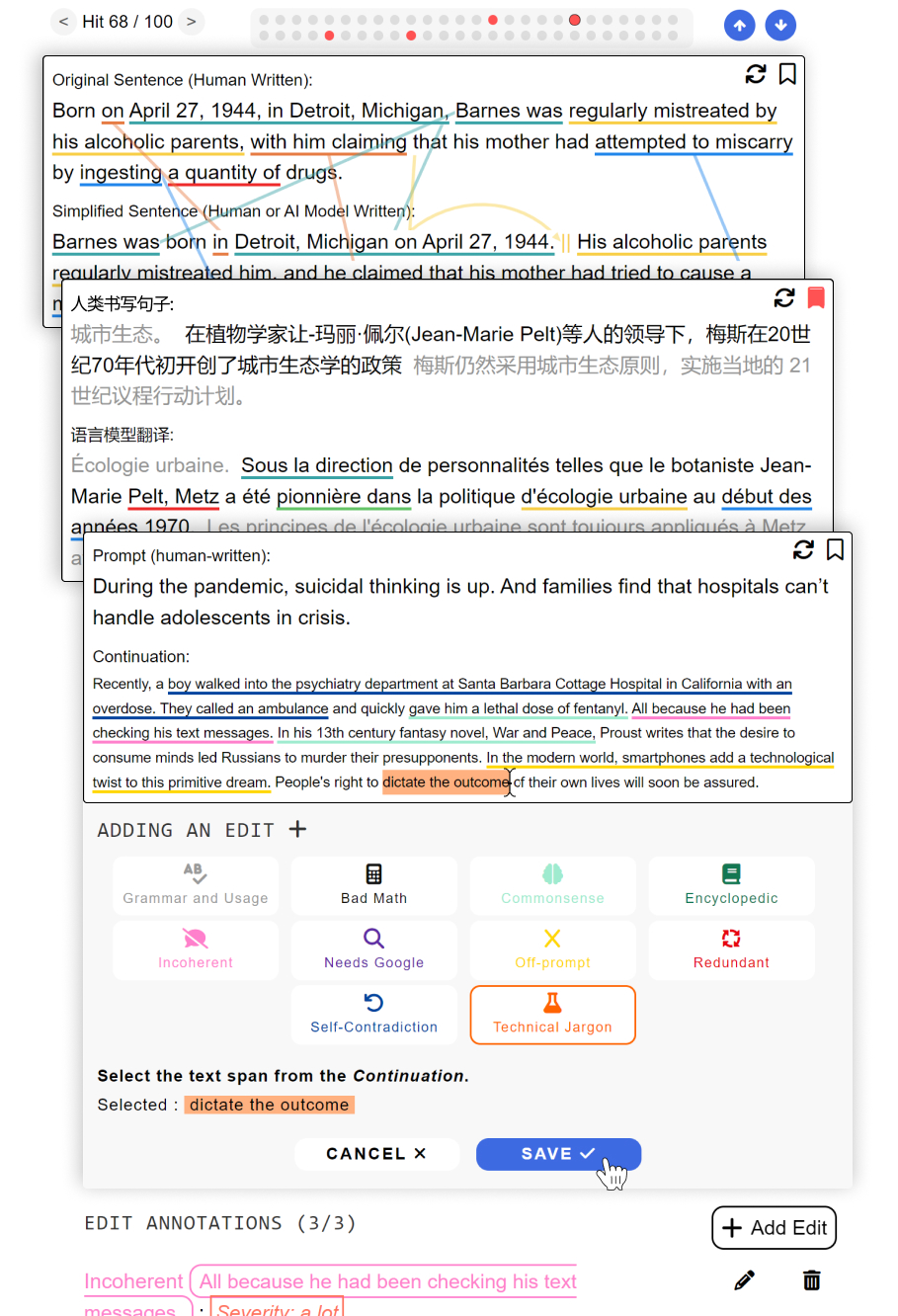}
    \setlength{\abovecaptionskip}{-3pt}
    \setlength{\belowcaptionskip}{-20pt}
    \captionof{figure}{Examples of fine-grained evaluation frameworks implemented on \platform{}. In order: SALSA \citep{heineman2023dancing}, MQM \citep{freitag-etal-2021-experts}, Scarecrow \citep{dou-etal-2022-gpt}.}
    \label{fig:overview}
\end{figure}

\setlength{\tabcolsep}{3pt}
\begin{table}[t!]
\centering
\renewcommand{\arraystretch}{1.0}
\resizebox{\columnwidth}{!}{
\begin{tabular}{llc} 
\toprule
\textbf{Framework} & \textbf{Task} & \textbf{Released} \\
\midrule
\textit{Evaluation} \\
MQM \citep{freitag-etal-2021-experts} & Translation & \cmark \\
FRANK \citep{pagnoni-etal-2021-understanding} & Summarization & \cmark \\
SNaC \citep{goyal-etal-2022-snac} & Narrative Summarization & \cmark \\
Scarecrow \citep{dou-etal-2022-gpt} & Open-ended Generation  & \cmark \\
SALSA \citep{heineman2023dancing} & Simplification & \cmark \\
ERRANT \citep{bryant-etal-2017-automatic} & Grammar Error Correction & \xmark \\
FG-RLHF \citep{wu2023fine} & Fine-Grained RLHF  & \cmark \\
\midrule
\textit{Inspection} \\
MultiPIT \citep{dou-etal-2022-improving} & Paraphrase Generation  & \xmark \\
CWZCC & Zamboanga Chavacano & \multirow{2}{*}{\xmark} \\
\citep{himoro-pareja-lora-2020-towards} & Spell Checking & \\
Propaganda & \multirow{2}{*}{Propaganda Analysis}  & \multirow{2}{*}{\cmark} \\
\citep{da-san-martino-etal-2019-fine} & & \\
arXivEdits \citep{jiang-etal-2022-arxivedits} & Scientific Text Revision & \cmark \\
\bottomrule
\end{tabular}}
\caption{Existing typologies currently implemented on \platform{}. \textit{Released} indicates whether the annotated data is released. Corresponding links on \platform{} for each framework can be found in Table \ref{tab:typologies_with_link} in the Appendix.}
\vspace{-14pt}
\setlength{\abovecaptionskip}{-4pt}
\label{tab:typologies}
\end{table}

However, each of these evaluation frameworks releases its own dedicated annotation interface that is difficult to modify or adapt to different evaluation schemes, thus limiting the customizablility. For example, Scarecrow's typology \cite{dou-etal-2022-gpt}, which is designed for open-ended text generation for news, may require modifications when applied to other domains such as story or scientific writing. Frameworks like MQM \cite{freitag-etal-2021-experts} only allow selections of the spans in the target sentence, restricting the ability to select the associated source spans in error categories such as mistranslation. Furthermore, modern LLMs are ideally evaluated on multiple tasks \cite{hendrycks2021measuring}, but the lack of a unified annotation tool makes this process inconvenient.
Considering the recent success of multi-task instruction fine-tuning \cite{wei2021finetuned,sanh2021multitask}, a standardized annotation format would enable research in multi-task learning with fine-grained human feedback.

To this end, we present \platform{} \logo{}: a unified and customizable platform for building, distributing and orchestrating fine-grained human evaluation for text generation in an efficient and easy-to-use manner. 
Our platform allows users to create, test and deploy an evaluation framework within minutes, all in a single browser window and has already been used to orchestrate large-scale data annotation \citep{heineman2023dancing}.
\platform{} also serves as a \textit{community hub} for fine-grained evaluation frameworks and annotation data, all presented in a unified format.
Figure \ref{fig:overview} displays three examples of evaluation frameworks built on \platform{}.
The following are the design principles of \platform{}:
\begin{itemize}[leftmargin=*,itemsep=0pt,topsep=2pt]
    \item \textbf{Unified:}
    \platform{} standardizes fine-grained evaluation into two key components: span selection and span annotation. Users can easily implement any framework by writing a YAML template file (see Figure \ref{fig:deployment}), and \platform{} will build the corresponding annotation interface. All resulting annotations adhere to a consistent JSON format.
    \item  \textbf{Customizable:} \platform{} offers extensive customization to meet a wide range of user needs. This includes different span selection methods from subword to word-level, diverse annotation options including custom questions and text boxes to handle arbitrary typologies, as well as customized interface elements in any language.
    \item  \textbf{Deployable:} \platform{} supports a range of deployment options for annotation projects of various scales. Small-scale linguistic inspections (e.g., manual ablation studies) can be directly hosted on the platform. For larger projects, users can host their template in a GitHub repository and connect to \platform{}. \platform{} is also compatible with crowdsourcing platforms such as Prolific\footnote{\url{https://www.prolific.co}} and Amazon MTurk\footnote{\url{https://www.mturk.com}}.
    \item \textbf{Contributive:} \platform{} also operates as a community hub where users can contribute and access a wide variety of fine-grained evaluation frameworks and their annotation data. Currently, it includes \ntypologies{} frameworks as displayed in Table \ref{tab:typologies}.
    \item \textbf{End-to-End:} Beyond facilitating the creation and deployment of evaluation frameworks, \platform{} streamlines every step of the annotation process. It offers functions for authors to publish their typologies as research artifacts and a supplementary Python library, released under the Apache 2.0 license, to help data collection.\footnote{\url{https://www.pypi.org/project/thresh}}
\end{itemize}

\section{Related Work}
\label{sec:related-work}
\noindent \textbf{Fine-grained Text Evaluation.}
Given the limitations of traditional human evaluation methods such as Likert-scale and pairwise comparison in the era of LLMs, many recent studies have proposed fine-grained human evaluation frameworks.
\citet{dou-etal-2022-gpt} introduces Scarecrow to capture error spans in open-ended text generation for news, MQM \cite{freitag-etal-2021-experts} identifies errors in machine translation, and FRANK
\cite{pagnoni-etal-2021-understanding} captures factual errors in abstractive text summarization.
We list other evaluation and inspection typologies in Table \ref{tab:typologies}.
However, these existing frameworks usually develop their own annotation tools which lack customizability and universality, making them difficult to adapt to other languages or domains, or to new annotation settings. 
Recently, \citet{goyal-etal-2022-falte} proposes FALTE, customizable span-level error highlighting for long text evaluation, but it only includes a subset of features offered by \platform{}, limiting its ability to implement complex typologies such as SALSA \cite{heineman2023dancing}. Specifically, FALTE only highlights errors without rating their severity or efficacy, does not support multi-span or composite selection, and cannot select overlapping spans. Moreover, its lack of a tree structure can make the interface cluttered if there are more than a handful of categories. \platform{} instead builds unified and customizable support across task setups.

\vspace{2pt}

\noindent \textbf{Annotation Tool.}
Accessible and replicable annotation tools have been a persistent goal for NLP tasks.
\citet{stenetorp-etal-2012-brat} introduces BRAT, the first web browser-based annotation tool and \citet{yimam-etal-2013-webanno} further improves BRAT on speed and label configuration. 
In recent years, a new generation of universal annotation tools have been introduced by academia and industry, including Prodigy \cite{prodigy_montani_honnibal}, Doccano \cite{doccano}, LightTag \cite{perry-2021-lighttag}, and POTATO \citep{pei-etal-2022-potato}.
Focusing on universality, these tools allow authors to add custom UI elements such as multiple choice questions, text boxes or pairwise comparison. 
However, these surface-level annotation options are not sufficient to implement complex typology setups demanded by fine-grained evaluation, which are typically structured by decision trees \cite{heineman2023dancing}. \platform{} addresses this gap by recursively building the interface, which allows for nested questions.
Besides, \platform{} encourages sharing and reproducibility by providing a community hub where users can upload their new or use existing fine-grained frameworks and annotated data.
\vspace{2pt}

\noindent \textbf{Span-level Annotation.}
Span-level annotation has a long history across NLP tasks.
In Named Entity Recognition (NER), spans are selected and labeled as names of persons, organizations, locations, or other entities \citep{tjong-kim-sang-de-meulder-2003-introduction}. 
Word alignment focuses on selecting aligned words or phrases between two parallel corpora across languages \citep{och-ney-2003-systematic}, or within monolingual tasks \citep{lan-etal-2021-neural}. 
Span selection has also been used for question answering such as in SQuAD \cite{rajpurkar-etal-2016-squad}, where the answer is defined by a span within the document context. 
Furthermore, extractive text summarization \citep{hermann2015teaching} highlights the spans that summarizes a given document. 
With a goal of understanding where and how text generation succeeds or fails, fine-grained text evaluation selects spans that are either quality or error in generated text.
These selected spans are then annotated following a complex typology and rated on the severity of errors or efficacy of high-quality content \cite{freitag-etal-2021-experts,dou-etal-2022-gpt,heineman2023dancing}.

\section{Fine-Grained Text Evaluation}
\label{sec:fine-grained-annotation}
\begin{figure}[t!]
    \centering
    \includegraphics[width=0.48\textwidth]{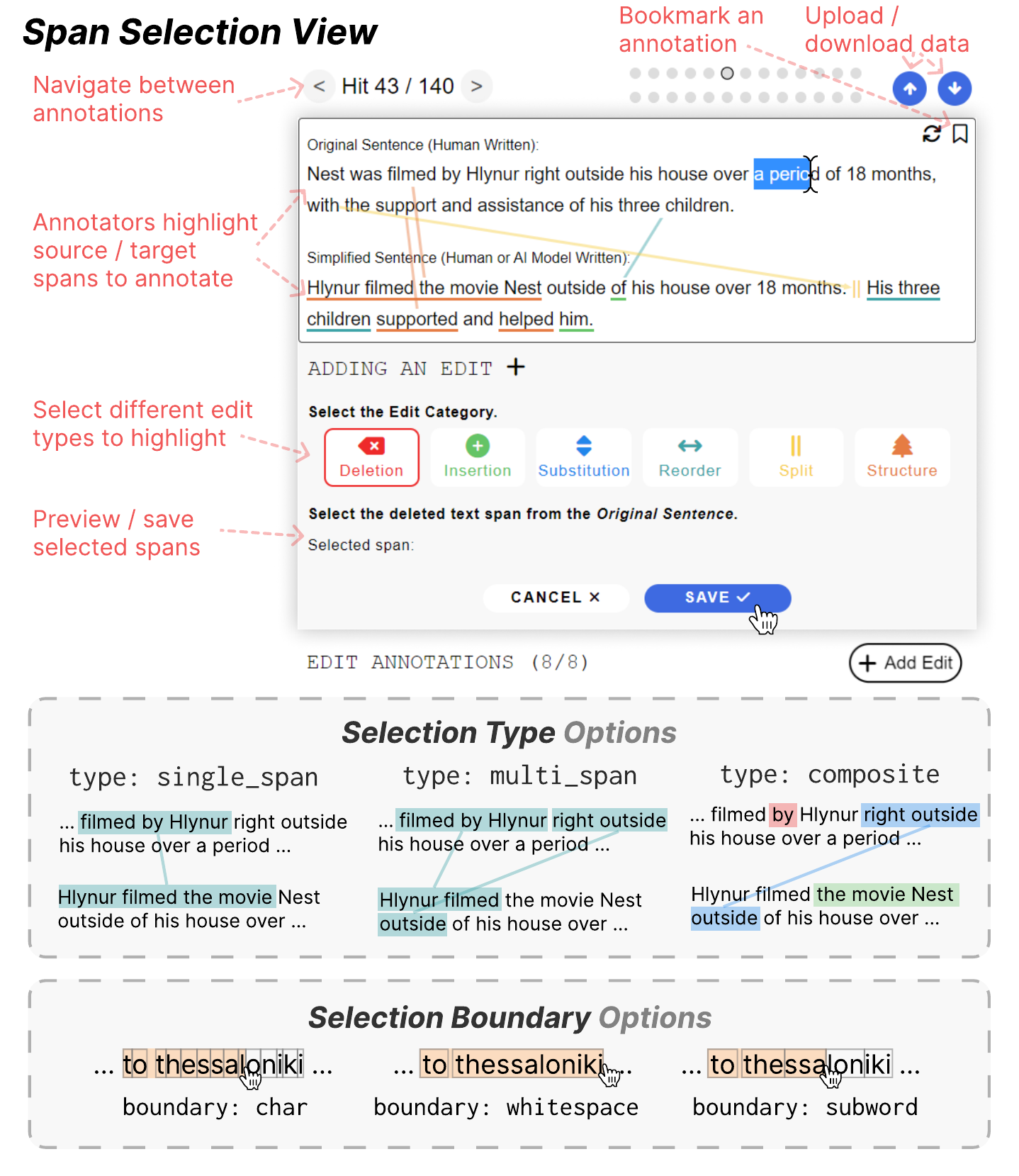}
    \setlength{\abovecaptionskip}{-14pt}
    \setlength{\belowcaptionskip}{-13pt}
    \captionof{figure}{The span selection component of \platform{}, customized with the SALSA \cite{heineman2023dancing} typology as an example.}
    \label{fig:span-selection-view}
\end{figure}

\begin{figure}[t!]
    \centering
    \includegraphics[width=0.48\textwidth]{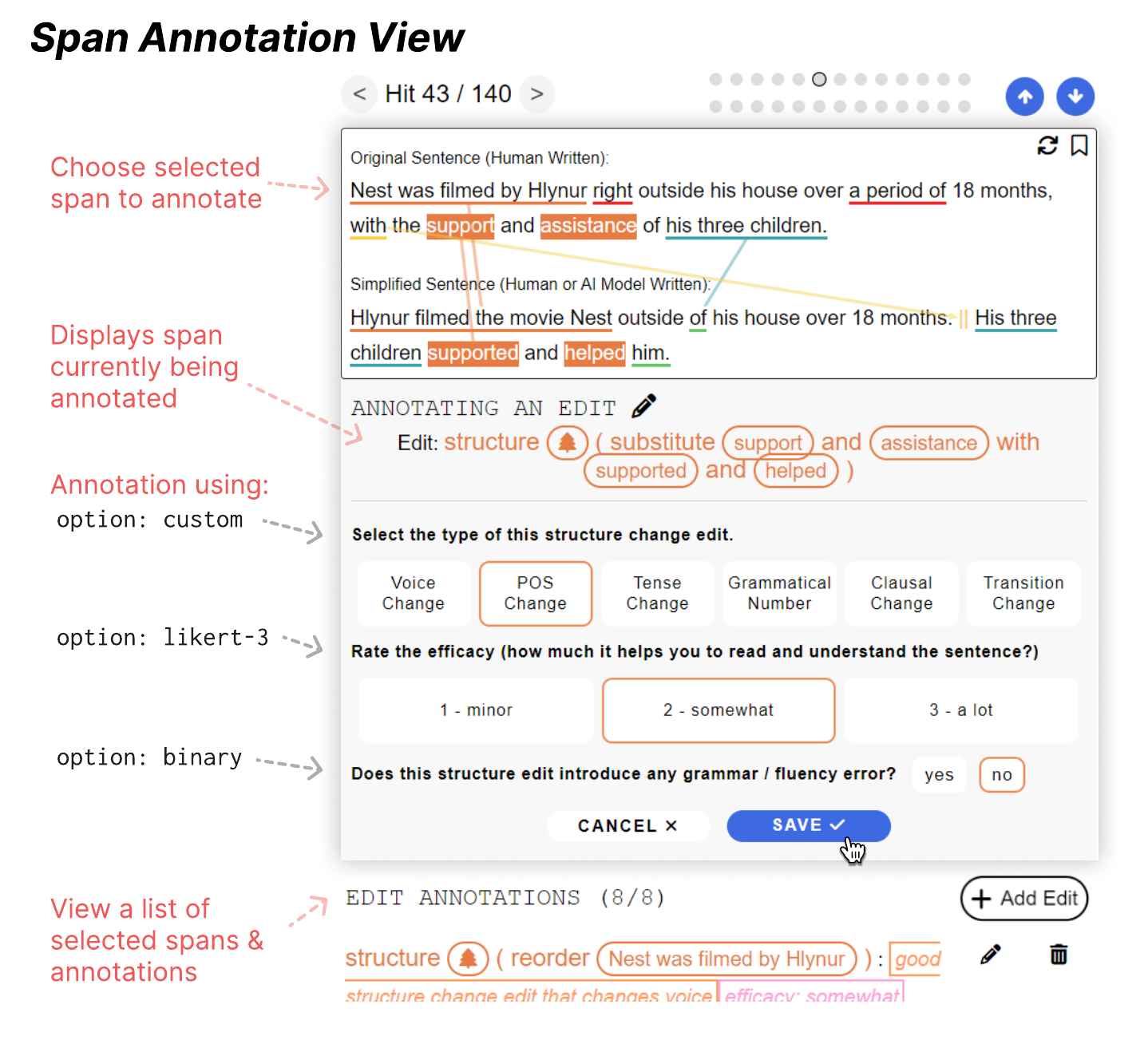}
    \setlength{\abovecaptionskip}{-14pt}
    \setlength{\belowcaptionskip}{-14pt}
    \captionof{figure}{The span annotation component of \platform{}, customized with the SALSA \cite{heineman2023dancing} typology as an example.}
    \label{fig:span-annotation-view}
\end{figure}

\platform{} formulates fine-grained text evaluation as two components: \textit{span selection} and \textit{span annotation}. During development, users define their annotation typology and interface features using a YAML template (see Sec \ref{sec:interface-builder} and Fig \ref{fig:deployment} for more details). Based on the configuration, \platform{} then constructs an annotation interface that integrates both components,
as illustrated in Figures \ref{fig:span-selection-view} and \ref{fig:span-annotation-view}.

\begin{figure*}[t!]
    \centering
    \includegraphics[width=\textwidth]{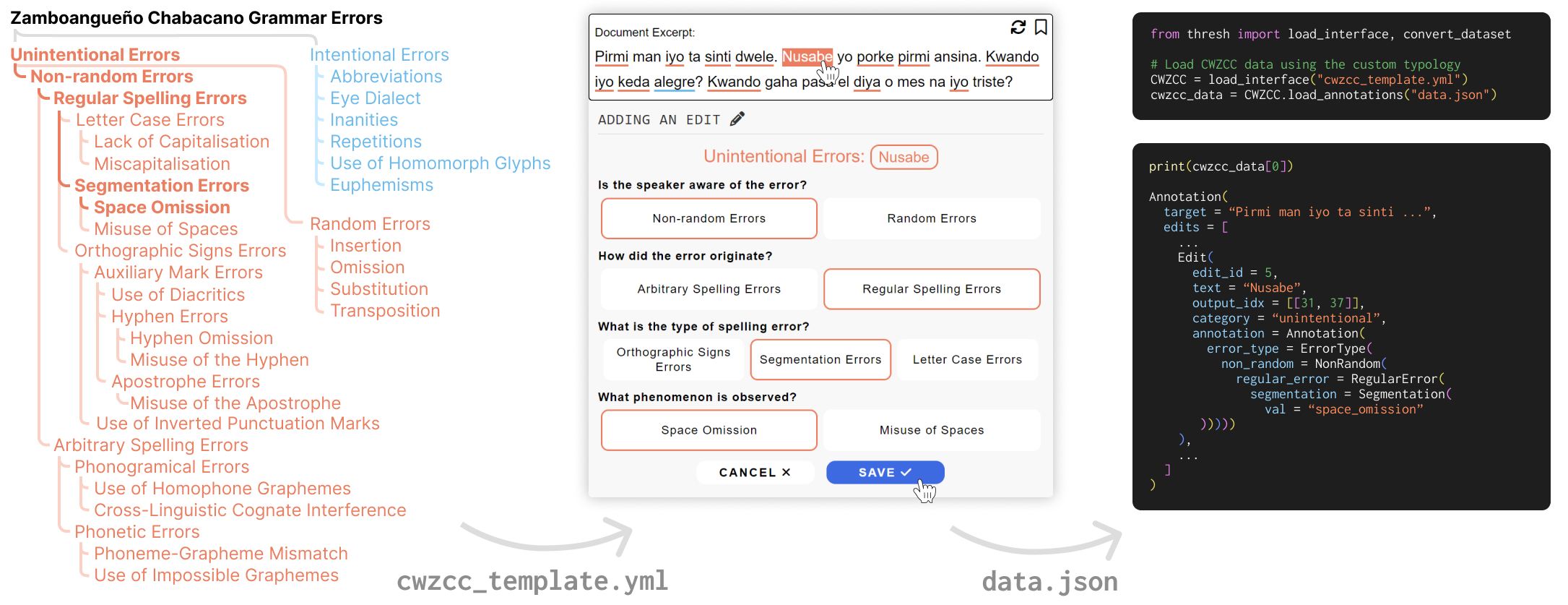}
    \setlength{\abovecaptionskip}{-10pt}
    \setlength{\belowcaptionskip}{-11pt}
    \captionof{figure}{The left figure shows a grammar error typology with 35 categories for contemporary written Zamboangueño Chabacano, a variant of Philippine Creole Spanish \cite{himoro-pareja-lora-2020-towards}. The center figure shows its annotation interface built on \platform{}, highlighting the ability for \platform{} to support complex, recursive annotation trees. The right figure shows the Python serialization for the annotation, generated by the \platform{} library.}
    \label{fig:annotation-tree}
\end{figure*}

\subsection{Span Selection}
Each annotation instance consists of the \textit{source}, \textit{target} and \textit{context}.
For example, in open-ended text generation \cite{zellers2019defending}, the source is a starting sentence and the target is a model-generated continuation. In text simplification \cite{xu-etal-2016-optimizing}, the source would be a complex sentence or paragraph, and the target would be the generated simplification.
The context holds additional relevant information, such as a prompt instruction, a retrieved Wikipedia page, or a dialogue history.
During the span selection stage, annotators select relevant spans, referred to as \textit{Edits}, in the source and target, following the edit category definitions outlined in the typology, as illustrated in Figure \ref{fig:span-selection-view}.
\vspace{2pt}

\noindent \textbf{Selection Type.}
For each edit category, users can specify one of three selection types: \textit{single-span}, \textit{multi-span}, or \textit{composite} -- the latter grouping together multiple single-span or multi-span selections. Multi-span selection is well-suited for edits that impact multiple parts of the source or target, e.g., the ``Redundant'' error in Scarecrow \cite{dou-etal-2022-gpt}, which requires selecting both the repetitive spans and their antecedents. Composite selections are ideal for high-level edits performed as a combination of several low-level edits, e.g., the ``Structure'' edit in SALSA \cite{heineman2023dancing}. Users can also customize each edit category to be selectable not only on the target, but also on the source (e.g., ``Deletion'' edit), or on both (e.g., ``Substitution'' edit), useful for text revision tasks.
\vspace{2pt}

\noindent \textbf{Selection Boundary.} Many span-selection interfaces define selection boundaries as each character, which can inadvertently lead to partial word selections and slow the annotation process. \citet{dou-etal-2022-gpt} proposes a solution that ``snaps'' the selection to the nearest whitespace, but this approach is limited in: (1) punctuation gets selected with adjacent words, even when this is not intended by annotators, (2) languages with no whitespace boundaries between words (e.g., Chinese) cannot be supported and (3) the annotation data cannot be perfectly translated to training data for token-level labeling tasks. We therefore introduce sub-word boundaries as a third option, in which users can use any LLMs tokenizer of their choice (such as \texttt{RobertaTokenizer} from Transformers\footnote{\url{https://www.github.com/huggingface/tokenizers}}) to tokenize the data and specify a \texttt{boundary: subword} flag in the YAML configuration file. 

\subsection{Span Annotation}
In the YAML file, users define the typology in a decision tree structure to further categorize the selected spans into fine-grained types. Unlike previous work which presents all fine-grained edit types to annotators simultaneously, \platform{} recursively compiles the annotation interface. Annotators thus will answer a series of questions or follow-up questions under each edit type, as shown in Figure \ref{fig:span-annotation-view}. This tree structure enables support for complex error typologies. An example of this can be seen in Figure \ref{fig:annotation-tree}, which shows a 35-category typology implementation for a grammar error correction task. \platform{} supports binary, three and five-scale questions with customized label names, as well as text boxes for tasks that require human post-editing or explanations. With these features, our interface supports complex annotation schemes in a flexible and easily extensible way.

We also give users the option of only enabling one of the two above components. This allows annotation for word/span alignment tasks \cite{sultan2014back} (where no annotation is needed) or two-stage annotation, where one set of annotators selects spans and then another set labels them.
\vspace{2pt}

\subsection{Additional Features}

\noindent \textbf{Adjudication View}. Using the \texttt{adjudication} flag, users can deploy two or three interfaces side-by-side, allowing adjudicators to inspect annotators' quality by comparing multiple candidate annotations simultaneously.
\vspace{2pt}

\noindent \textbf{Multi-Language Support.} Fine-grained evaluation has seen almost exclusive attention to English tasks \citep{huidrom-belz-2022-survey}. To smoothen the deployment barrier for multilingual fine-grained evaluation, all interface elements can be overridden to suit any language. For our default interface text, we support \ntranslations{} translations which can be enabled out-of-the-box by adding a \texttt{language} flag: \textit{zh}, \textit{en}, \textit{es}, \textit{hi}, \textit{pt}, \textit{bn}, \textit{ru}, \textit{ja}, \textit{vi}, \textit{tr}, \textit{ko}, \textit{fr} and \textit{ur}.
\vspace{2pt}

\noindent \textbf{Instructions.} Users may write interface instructions with Markdown formatting, which allows for links, pictures and inline code. They have the option to display their instructions as a pop-up modal, or prepend the text above the interface.
\vspace{2pt}

\noindent \textbf{Paragraph-level Annotation.} By breaking evaluation down to individual sentences, authors can reduce the cognitive load required for lengthy annotation tasks such as identifying errors in long-form summarization \citep{goyal-etal-2022-falte}. Users can specify an additional \texttt{context\_before} or \texttt{context\_after} field to add paragraph-level context or custom \texttt{display} options to view paragraphs text side-by-side with selected edits.

\section{Interactive Interface Builder}
\label{sec:interface-builder}
\begin{figure*}[th]
    \centering
    \includegraphics[width=\textwidth]{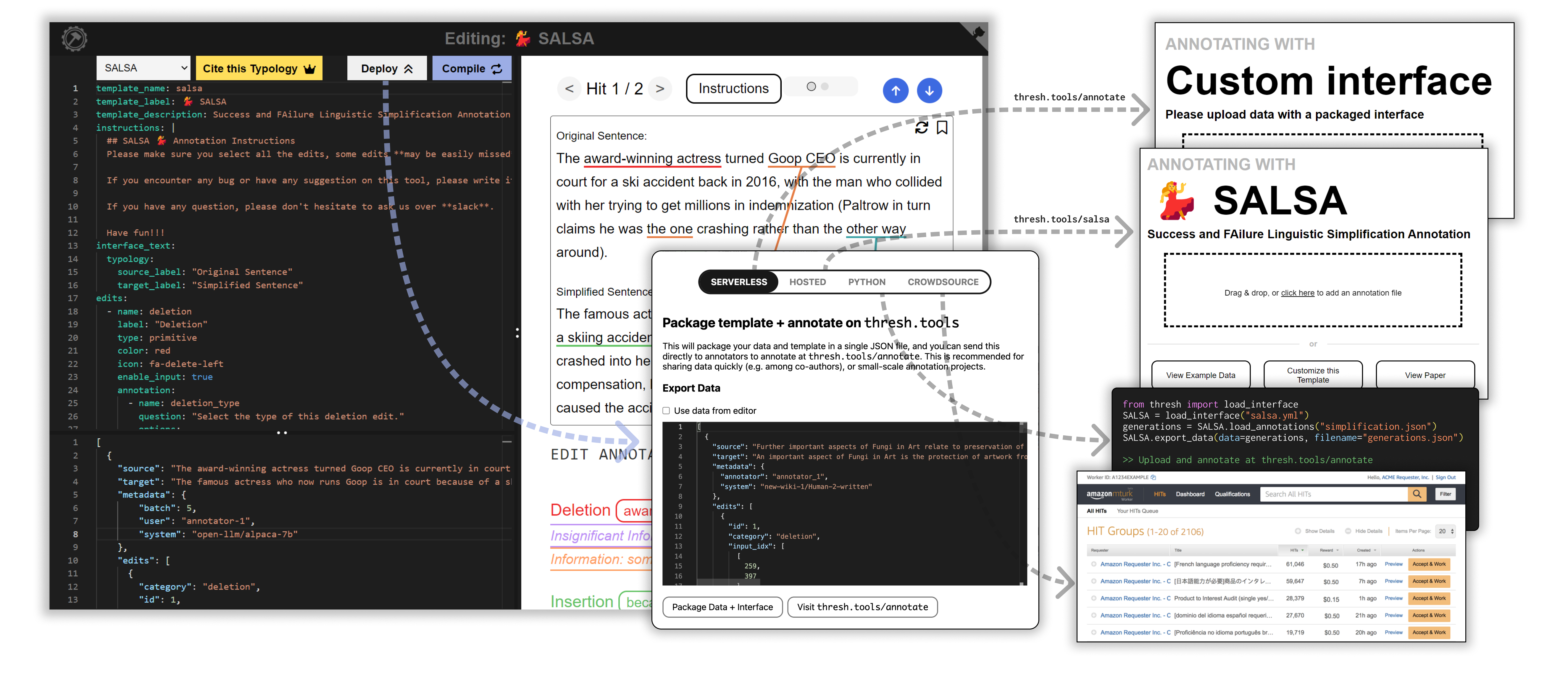}
    \setlength{\abovecaptionskip}{-15pt}
    \setlength{\belowcaptionskip}{-13pt}
    \captionof{figure}{\platform{} deployment workflow. Users build and test their template and then deploy with one of 4 options.}
    \label{fig:deployment}
\end{figure*}

To alleviate the time consuming process of customizing and hosting front-end code — even building custom databases in some cases — \platform{} implements an in-browser interface builder, which allows users to create, test and deploy a fine-grained interface within a single web browser page, as depicted in Figure \ref{fig:deployment}.
Users write a YAML template to construct their interface and provide data with a JSON file. The \textit{Compile} button allows users to preview their interface, and the \textit{Deploy} button presents instructions for different deployment options, which are described in \S\ref{sec:deployment}.
\vspace{2pt}

\noindent \textbf{Template Hub.} As \platform{} aims to facilitate easy use and distribution of fine-grained evaluation frameworks, it provides a template hub that makes it simple for any NLP practitioner to access a framework with their own data. Alongside the \ntutorials{} tutorial templates that explain each interface feature, the annotation builder currently includes \ntypologies{} widely used inspection and evaluation typologies across major text generation tasks. Table \ref{tab:typologies} (on Page 2) lists each framework, its associated task and link to our implementation.

To upload a framework to \platform{}, users can create a GitHub pull request with their typology's YAML file, which is merged publicly. We also include other features to facilitate sharing and replication. Users can add a \texttt{citation} flag along with a BibTex citation, which creates a \textit{Cite this Typology} button in the annotation builder, a \texttt{paper\_link} flag,  which adds a link to their research paper in the builder and on deployment, and a \texttt{demo\_data\_link} flag which creates a \textit{View Demo Data} button to allow viewers to use the interface with example data.

For testing, users can paste data into the interface builder interactively, and for deployment can link to data files. Data can be blank or come with existing annotations, in which case the annotations will be appropriately parsed, verified and rendered.
\vspace{2pt}

\noindent \textbf{Unified Data Model.} As shown in Table \ref{tab:typologies} on Page 2, many existing frameworks have released their annotated data, but in varied formats. To ensure compatibility, we create conversion scripts that adapt these annotations to our unified format. Our scripts are designed to be \textit{bidirectional}, meaning data published for these typologies can be converted to our format and back without data loss. Our unified fine-grained data format allows smooth transfer of analysis, agreement calculation and modeling code between different projects.
We believe this will support research in learning with multi-task fine-grained training setups or model feedback.
Like framework templates, users can upload their annotated data to the hub via a GitHub pull request.

\section{Deployment}
\label{sec:deployment}
Managing and collecting fine-grained annotations becomes bulky at scale, we thus release supplementary tools to deploy interfaces quickly or programmatically, and integrate loading annotations directly into Python. This includes the \pythonlib{} library\footnote{\url{https://www.pypi.org/project/thresh}}, which is useful for compiling interfaces and loading annotations. We support the following deployment types as shown in Figure \ref{fig:deployment}:

\begin{itemize}[leftmargin=*,itemsep=0pt,topsep=2pt]
\item \noindent\textbf{Hosted:} Best for small-scale inspection or data exploration, users can download a file that bundles the data and template together. Then, users can upload this file to \url{thresh.tools/annotate} to begin annotation immediately.

\item \noindent\textbf{Serverless:} Users upload their YAML template to a public repository such as GitHub or HuggingFace, and link their template to \texttt{thresh.tools} through a URL parameter: \texttt{gh} or \texttt{hf} respectively. Users can also link data via the \texttt{d} parameter. In addition, we release demo code for users to host their interface on their own domain without cloning the \platform{} repository.

\item \noindent\textbf{Python:} For large scale projects, users can programmatically generate and deploy templates using the \texttt{create\string_template} functionality provided in the \pythonlib{} library. This helps for projects with a large number of templates, such as annotation in multiple languages. Additionally, integration with Python allows a direct connection from model generation to annotation processing, supporting the creation of workflows like fine-grained RLHF \citep{wu2023fine}.

\item \noindent\textbf{Crowdsource:} If the data collection process is mishandled, annotation by crowdworkers can lead to poorly standardized or noisy data \citep{karpinska-etal-2021-perils,veselovsky2023artificial}. To assist annotation quality control, we publish tools to encourage best practices when using crowdsource platforms. Our crowdsource deployment workflow includes example code for interactive, multi-stage tutorials to create qualification tasks and step-by-step tutorials for deployment on both Prolific and Amazon Mechanical Turk.
\end{itemize}

\noindent Additionally, we support lightweight database integration (such as with Google Firebase\footnote{\url{https://firebase.google.com}}) for all deployment types, allowing users to connect their own database to any annotation setup.
\vspace{2pt}

\noindent \textbf{Python Serialization.} Compared to previous work that simply exports JSON annotations, our supplementary \pythonlib{} library includes functionality for loading and combining annotation files to simplify the data ingestion process. For example, \texttt{load\string_annotations} merges multiple data files, serializes the data into Python objects, and evaluates whether the data collected is consistent with the configuration used to load the data.

\section{Conclusion}
\label{sec:conclusion}
We present \platform{} \logo{}, a unified, customizable, and deployable platform for fine-grained text evaluation.
\platform{} offers extensive customization via a simple YAML configuration file, and facilitates a community hub for sharing frameworks and annotations.
The platform also ensures seamless deployment for any scale of annotation projects and introduces a Python library to further ease the process from typology design to annotation processing.

\section*{Ethical Considerations}
We do not anticipate any ethical issues pertaining to the topics of fine-grained evaluation supported by our interface. Nevertheless, as \platform{} lowers the barrier to fine-grained evaluation, vast ethical responsibility falls upon practitioners using our platform to prevent the exploitation of crowdsource workers, through fair pay \citep{fort-etal-2011-last} and safeguards against exposure to harmful or unethical content \citep{shmueli-etal-2021-beyond}. As task difficulty and complexity scales with the granularity of data collected, increasing care must be taken for training annotators adequately and to scale pay accordingly \citep{williams2019perpetual}.

\section*{Acknowledgments}
This research is supported in part by the NSF awards IIS-2144493 and IIS-2112633, ODNI and IARPA via the HIATUS program (contract 2022-22072200004). The views and conclusions contained herein are those of the authors and should not be interpreted as necessarily representing the official policies, either expressed or implied, of NSF, ODNI, IARPA, or the U.S. Government. The U.S. Government is authorized to reproduce and distribute reprints for governmental purposes notwithstanding any copyright annotation therein.

\bibliography{anthology,custom}
\bibliographystyle{acl_natbib}

\appendix
\clearpage
\begin{table*}[t!]
\centering
\small

\renewcommand{\arraystretch}{1.1}

\resizebox{\textwidth}{!}{
\begin{tabular}{llcl}
\toprule
\textbf{Framework} & \textbf{Task} & \textbf{Released} & \textbf{Link} \\
\midrule
\textit{Evaluation} \\
MQM \citep{freitag-etal-2021-experts} & Translation & \cmark & \texttt{\href{https://thresh.tools/mqm}{thresh.tools/mqm}} \\
FRANK \citep{pagnoni-etal-2021-understanding} & Summarization & \cmark & \texttt{\href{https://thresh.tools/frank}{thresh.tools/frank}} \\
SNaC \citep{goyal-etal-2022-snac} & Narrative Summarization & \cmark & \texttt{\href{https://thresh.tools/snac}{thresh.tools/snac}} \\
Scarecrow \citep{dou-etal-2022-gpt} & Open-ended Generation  & \cmark & \texttt{\href{https://thresh.tools/scarecrow}{thresh.tools/scarecrow}} \\
SALSA \citep{heineman2023dancing} & Simplification & \cmark & \texttt{\href{https://thresh.tools/salsa}{thresh.tools/salsa}} \\
ERRANT \citep{bryant-etal-2017-automatic} & Grammar Error Correction & \xmark & \texttt{\href{https://thresh.tools/errant}{thresh.tools/errant}} \\
FG-RLHF \citep{wu2023fine} & Fine-Grained RLHF  & \cmark & \texttt{\href{https://thresh.tools/fg-rlhf}{thresh.tools/fg-rlhf}} \\
\midrule
\textit{Inspection} \\
MultiPIT \citep{dou-etal-2022-improving} & Paraphrase Generation  & \xmark & \texttt{\href{https://thresh.tools/multipit}{thresh.tools/multipit}} \\
CWZCC \citep{himoro-pareja-lora-2020-towards} & Zamboanga Chavacano Spell Checking & \xmark & \texttt{\href{https://thresh.tools/cwzcc}{thresh.tools/cwzcc}} \\
Propaganda \citep{da-san-martino-etal-2019-fine} & Propaganda Analysis  & \cmark & \texttt{\href{https://thresh.tools/propaganda}{thresh.tools/propaganda}} \\
arXivEdits \citep{jiang-etal-2022-arxivedits} & Scientific Text Revision & \cmark & \texttt{\href{https://thresh.tools/arxivedits}{thresh.tools/arxivedits}} \\
\bottomrule
\end{tabular}}
\caption{Existing typologies implemented on \platform{} with their associated link. \textit{Released} indicates whether the annotated data is released.}
\setlength{\belowcaptionskip}{-5pt}
\label{tab:typologies_with_link}
\end{table*}

\end{document}